\documentclass[letterpaper, 10 pt, conference]{ieeeconf}

\IEEEoverridecommandlockouts

\usepackage{dblfloatfix}
\usepackage[noadjust]{cite}
\usepackage{amsmath,amssymb,amsfonts}
\usepackage{algorithmic}
\usepackage{graphicx}
\usepackage{textcomp}
\usepackage{xcolor}
\usepackage{pifont}
\usepackage{multicol}
\usepackage{multirow}
\usepackage{tabularx}
\usepackage[caption=false,font=footnotesize]{subfig}
\usepackage[citecolor=blue, colorlinks=true, urlcolor=black, linkcolor=red]{hyperref}
\usepackage{algorithm}
\usepackage{algorithmic}

\DeclareMathOperator*{\argmin}{\arg\!\min}

\begin{document}

\title{\LARGE Visual Inertial Odometry using Focal Plane Binary Features (BIT-VIO)}

\author{Matthew Lisondra$^{\dagger\star}$, 
        Junseo Kim$^{\dagger\star}$, 
        Riku Murai$^\ddagger$, 
        Kourosh Zareinia$^\dagger$, 
        and Sajad Saeedi$^\dagger$%
\thanks{
$^\dagger$Toronto Metropolitan University (TMU).}
\thanks{
$^\ddagger$Imperial College London, Department of Computing.}
\thanks{*Both authors contributed equally to this research at TMU.}
}

\maketitle

\begin{abstract}

Focal-Plane Sensor-Processor Arrays (FPSP)s are an emerging technology that can execute vision algorithms directly on the image sensor. Unlike conventional cameras, FPSPs perform computation on the image plane -- at individual pixels -- enabling high frame rate image processing while consuming low power, making them ideal for mobile robotics. 
FPSPs, such as the SCAMP-5, use parallel processing and are based on the Single Instruction Multiple Data (SIMD) paradigm. 
{In this paper, we present BIT-VIO, the first Visual Inertial Odometry (VIO) which utilises SCAMP-5.}
{BIT-VIO is a loosely-coupled iterated Extended Kalman Filter (iEKF) which fuses together the visual odometry running fast at 300 FPS with predictions from 400 Hz IMU measurements to provide accurate and smooth trajectories.\\
Project Page: \url{https://sites.google.com/view/bit-vio/home}
}

\end{abstract}

\IEEEpeerreviewmaketitle

\section{Introduction}
{The reduced power consumption and latency associated with Visual Odometry (VO) and Visual Inertial Odometry (VIO) are becoming increasingly important as future mobile devices are anticipated to require rich and accurate spatial understanding capabilities~\cite{ieeep_2018}.
Currently, conventional camera technology typically operates at 30-60 frames per second (FPS) and transfers a non-trivial amount of data from the sensor to the host device (e.g. a desktop PC).
Such data transfer is not free -- in terms of both power and latency --, and additionally, all these pixels must be then later processed on the host device.
}

{As an alternative, Focal-Plane Sensor-Processor Arrays (FPSP)s, such as SCAMP-5, is a new technology that enables computation to occur on the imager's focal plane before transferring the data to a host-device~\cite{dudek2011scamp}. 
By performing early-stage computer vision algorithms on the focal plane such as feature detections, FPSPs compress the image data down to the size of the features. By transferring only the detected features, redundant pixel information is not transferred or potentially even not digitized as FPSPs such as SCAMP-5 can perform analog computation.}

\begin{figure}
\centerline{\includegraphics[width = 0.4 \textwidth]{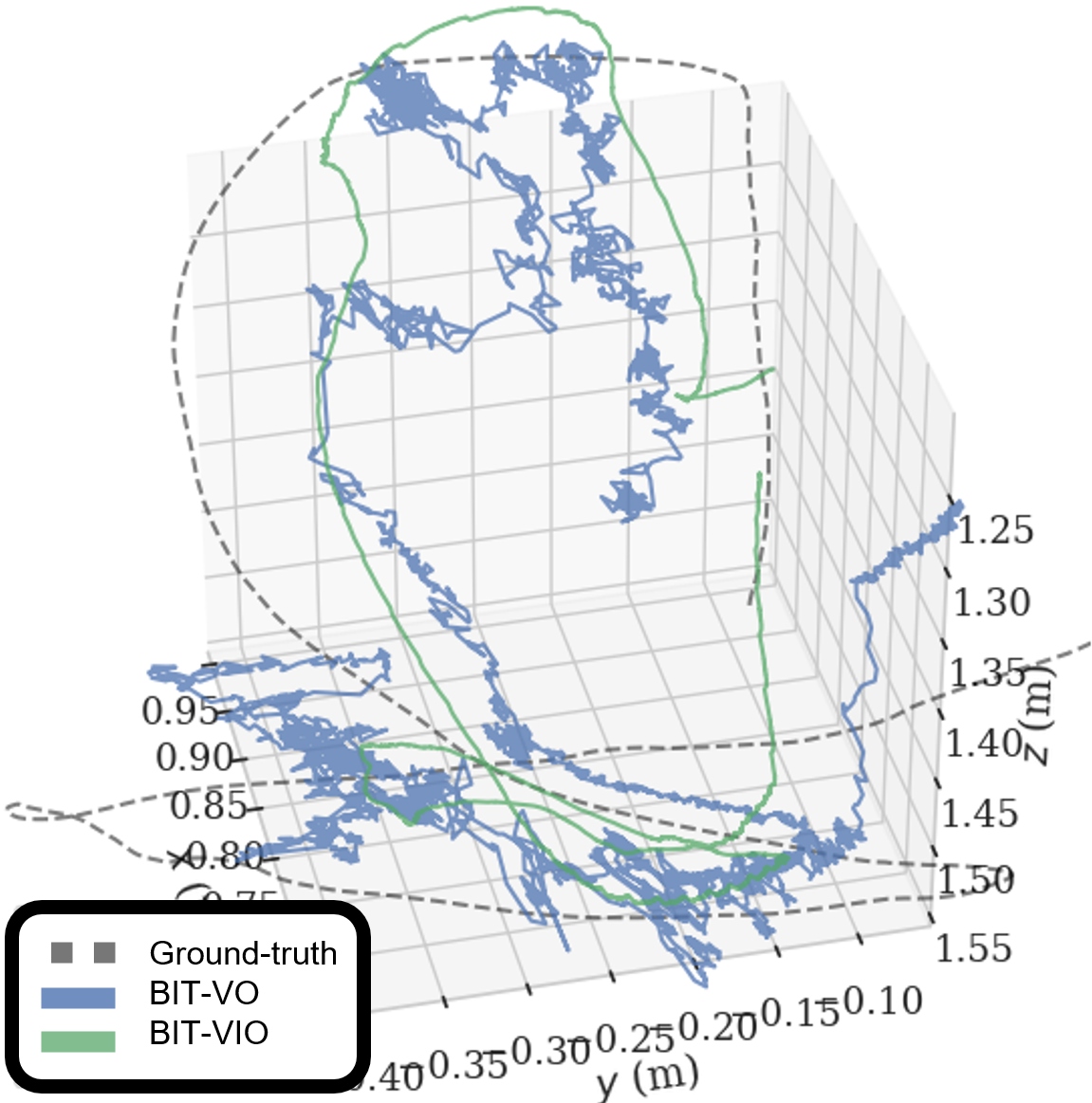}}
\caption{
\color{black} Comparison of  \color{black} the proposed BIT-VIO algorithm \color{black} and visual odometry (BIT-VO) overlaid on the reference ground-truth trajectory. BIT-VIO estimates are closer to the ground-truth trajectory compared to predictions from BIT-VO. Notice that BIT-VIO effectively removes the high-frequency noise visible in BIT-VO's trajectory. The plot was generated using evo~\cite{grupp2017evo}. \color{black}
}
\label{fig:BIT-VIO-100-pt2}
\label{fig}
\end{figure}

{In this work, we extend on BIT-VO~\cite{murai2020bit, murai2023bit}, a visual odometry algorithm which uses SCAMP-5, and present BIT-VIO, the first 6-Degrees of Freedom (6-DOF) Visual Inertial Odometry (VIO) algorithm to utilize the advantages of the FPSP for vision-IMU-fused state estimation. As shown in Fig.~\ref{fig:BIT-VIO-100-pt2}, BIT-VIO achieves a much smoother trajectory estimate when compared to BIT-VO, while retaining all the advantageous properties of BIT-VO such as low latency and high frame rate pose estimation.}
In short, the contributions of our work are:
\begin{itemize}
    \item Efficient Visual Inertial Odometry operating and correcting by loosely-coupled sensor-fusion iterated Extended Kalman Filter (iEKF) at 300 FPS using predictions from IMU measurements obtained at 400 Hz. 
    \item Uncertainty propagation for BIT-VO's pose as it is based on binary-edge-based descriptor extraction, 2D to 3D re-projection.
    \item Extensive real-world comparison against BIT-VO, with ground-truth obtained using a motion capture system. 
\end{itemize}

The remainder of this work is organized as follows.
Sec.~\ref{sec:background} describes the background about SCAMP-5 FPSP. 
Sec.~\ref{sec:method} explains the proposed BIT-VIO algorithm. 
Sec.~\ref{sec:results} details our experimental results. Finally, Sec.~\ref{sec:conc} concludes the work.

\section{Background}
\label{sec:background}
In this section, we review SCAMP-5, an FPSP technology developed by the University of Manchester~\cite{dudek2011scamp},~\cite{chen2018scamp5d}, and it's application to robotics and visual odometry.

\subsection{SCAMP-5 FPSP}
SCAMP-5 is a $256\times256$ processor array, performing parallel processing in a SIMD fashion on the focal plane of the imaging sensor. The parallelism of the SCAMP-5 FPSP camera technology provides high computational capabilities and the on-sensor processing enables low-power operation. 
Each pixel has a processing element (PE) that contains $7$ analog registers, $13$ digital registers, and an ALU, enabling on-pixel logical and arithmetic operations. 
The analog registers can store a real-valued variable, 
up to around $8$-bit resolution. 
The analog registers can do operations such as add, negate, split, and compare-against-$0$. 
The digital registers can do MOV, OR, NOR, and NOT operations. Each PE can communicate EAST, WEST, NORTH, and SOUTH with its neighboring PEs analog registers and digital registers.

Beyond SIMD parallelism, SCAMP-5's digital registers can be read out as events. Event readout only transfers coordinates of the digital registers set to $1$ and is more efficient than transferring the whole image plane if only a sparse set 
and pixels have a register set to $1$~\cite{chen2018scamp5d}. 
Flooding is a feature of the SCAMP-5 that enables DREG propagation of $1$s to help further accelerate image processing on its hardware. Through the SCAMP-5 camera technology's asynchronous propagation network, the flooding speed is nearly $62\times$ faster than accessing the neighbouring pixels on the SCAMP-5 via conventional message passing~\cite{carey2013100}.


\subsection{Application of SCAMP-5 to Robotics}
The SCAMP-5 has been utilized across many robotic systems.  
The odometry system by Greatwood et al. performs HDR edge-based odometry on the SCAMP-5~\cite{greatwood2018perspective} and achieves lower power consumption than a conventional camera system. Additionally, the high frame rate nature of the system meant that it suffered less from motion blurs under agile motions. In \cite{boseVisualOdometryPixel2017}, SCAMP-5 is used to track the 4 DoF camera motions using direct image alignment, and all computation is performed on the sensor itself. In \cite{fourdofcamtracking}, another algorithm is proposed, performing optical flow to estimate 4 DoF camera motion. 
In~\cite{mcconville2020visual}, SCAMP-5 is utilized for visual odometry on unmanned UAVs. Compared to using a conventional camera, they demonstrated that SCAMP-5-based systems have a clear practical advantage, for example, by computing HDR on the SCAMP-5, UAVs can transition from outdoor to indoor environments whilst successfully tracking, despite the changes in the lighting conditions.

The high frame-rate nature of SCAMP-5 also opens up the possibility for many interesting applications. For example, 
in-sensor CNN inference can perform hand gesture recognition for a rock, paper, scissors game at 8000 FPS, and can always make a robot play a winning hand~\cite{liu2021direct}. While it is a simple setup, the game requires the system to have low end-to-end latency and is challenging to replicate using a conventional camera.

Fully utilising the different computational capabilities available on the SCAMP-5 device,~\cite{castillo2021weighted} 
performs all-on-sensor mapping and localization. It performs visual route mapping and localization on the SCAMP-5 and runs at more than 300 FPS on various large-scale datasets.

SCAMP-5 has also been applied for controls, for example, using focal-plane processing, a ground target was detected and tracked to guide a small, agile quadrotor UAV~\cite{greatwood2017tracking}.
In~\cite{greatwood2019towards}, they perform drone racing, using the SCAMP-5 to detect the gates. The gate size and location are the only data that is transferred, with minimal data transfer resulting in 500 FPS. 
In~\cite{chen2020proximity}, different visual features such as corner points, blobs, and edges are extracted on the SCAMP-5 and fed into a recurrent neural network (RNN) for obstacle avoidance.
In-sensor analog convolutions have been proposed in \cite{Wong2020AnalogNetCN} and \cite{auke}.
AnalogNavNet~\cite{stow2022compiling} utilized Cain~\cite{stow2020cain} to implement a CNN that operates on the analog registers on SCAMP-5 for robotic navigation inside a corridor and racetrack environment.

\subsection{BIT-VO}
Our method builds on the previous work BIT-VO~\cite{murai2020bit}, which performs 6 degrees of freedom (DoF) visual odometry at 300 FPS using a SCAMP-5 camera.
In BIT-VO, the VO system is clearly separated into a frontend, which performs feature extraction, and a backend, which performs the matching of the features and the camera pose optimization.
Following this separation, BIT-VO performs the frontend feature detection on the SCAMP-5 camera itself, where corners and binary edges are detected and transferred at 300 FPS utilising the SIMD processing capability of SCAMP-5 and the event readouts.

The detected features are then transferred to a host device, which performs the backend processing. Here, for each corner feature, a descriptor is formed using the binary edges. Using brute form matching, corners across frames are matched using the descriptors, similarly to ORB-SLAM~\cite{mur2015orb}.
Once the correspondences are established, the system is initialised using a 5-point algorithm~\cite{nister2004efficient} and after the initialization, the camera pose is optimized by minimising the map-to-frame reprojection error.

By operating at 300 FPS, BIT-VO is robust against rapid, agile camera motions. However, the estimated trajectory contains a high-frequency noise, which is due to the noisy feature detection on the focal plane. In this work, we aim to address this problem by incorporating IMU measurements.

\subsection{Visual Inertial Odometry}
Visual inertial odometry (VIO) is the process of estimating camera pose by combining visual information from a camera and inertial measurements from IMUs.
VIO provides accurate and robust pose estimates. The sensors complement each other and are used in many applications and products such as AR headsets.
VIO can be categorised into loosely-coupled and tightly-coupled methods. In a loosely-coupled method, the visual and inertial measurements are independently processed to estimate the motion and then are fused together for correction.
On the other hand, the tightly-coupled method directly estimates the motion from the visual and inertial measurements~\cite{scaramuzza2020aerial}.

There are many different ways one can implement a VIO system. For example, one can use filtering~\cite{mourikis2007multi, bloesch2015robust} approaches, or using non-linear optimization, perform fixed-lag smoothing~\cite{leutenegger2015keyframe, qin2018vins} or even full smoothing~\cite{forster2016manifold}.
For the full smoothing, solving the entire system at every observation quickly becomes infeasible, hence they rely on iSAM2~\cite{kaess2012isam2} for incremental factor graph optimization.

\subsection{Visual Inertial Odometry on Unconventional Cameras}
While a conventional camera is typically used in VO and visual SLAM, other camera technologies such as event-based cameras are used in many state-of-the-art VIO algorithms~\cite{zihao2017event},~\cite{rebecq2017real},~\cite{vidal2018ultimate},~\cite{mueggler2018continuous}.
Event cameras provide low power usage and low latency benefits over conventional cameras and are also robust against illumination changes~\cite{lichtsteiner2008128}. However, while
event cameras compress visual information into a continuous stream of events, they are not user-programmable and cannot extract a specific feature such as FAST-corner on the sensor itself~\cite{rosten2006machine, Chen:etal:IROS2017}. Furthermore, the data volume transferred by an event camera is proportional to camera motion, and such a characteristic is not optimal; for instance, a robot has more data to process during rapid motion.

\section{Proposed Method}
\label{sec:method}
In this section, we first introduce the notations and then present an overview of the whole system.

\subsection{Notations}
\begin{figure}[t]
\centerline{\includegraphics[width = 0.4 \textwidth]{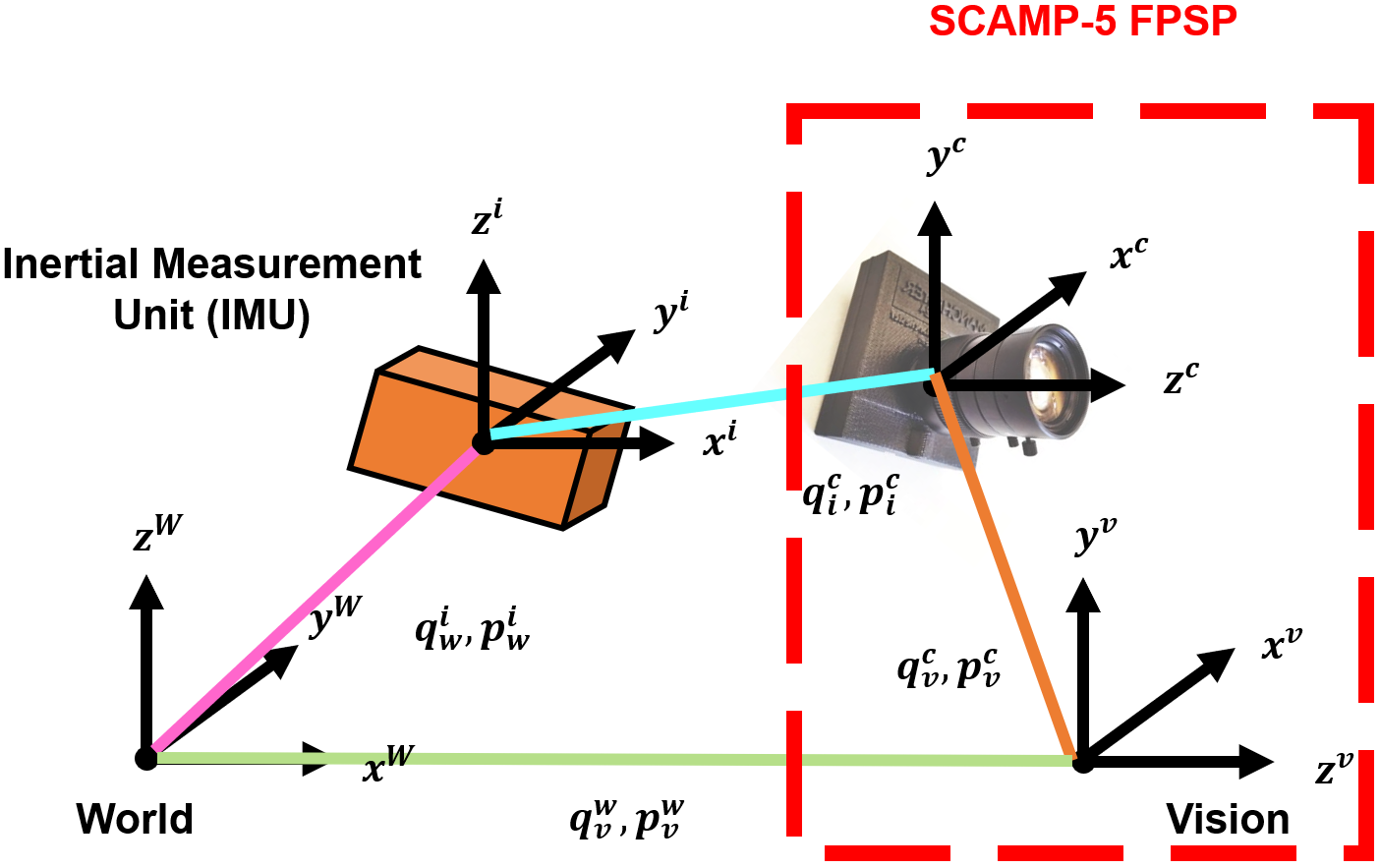}}
\caption{Coordinate frame definition of the IMU and the SCAMP-5.  In total, we define four coordinate frames. Notation \(p_{A}^{B}\) and \(q_{A}^{B}\) are used to represent transformation from $A$ to $B$. \color{black}}
\label{fig:frames}
\end{figure}

The following notation conventions are used in this work, adopted from~\cite{weiss2011real},~\cite{trawny2005indirect}:

\begin{itemize}
    \item Units of a variable \(A\) as \([A]\) (e.g. \([a_x] = m/s^2\)).
    \item Skew-symmetric matrix of \(A\) is \(\lfloor A \rfloor\).
    \item \(p_{A}^{B}\) represents the translation from frames \(A \rightarrow B\).
    \item \(q_{A}^{B}\) represents the Hamiltonian quaternion rotation \( ({q_{A}^B}_w , {q_{A}^B}_x, {q_{A}^B}_y, {q_{A}^B}_z   ) \) from frames \(A \rightarrow B\).
    \item \(\hat{p}, \hat{q}\) are the expected translation and rotation.
    \item \(\tilde{p}, \tilde{q}\) are the error in translation and rotation.
    \item \(C_{(q)}\) is the rotational matrix to the quaternion \(q\).
    \item \(\Omega(\omega)\) is quaternion-multiplication matrix of \(\omega\).
    \item \(\delta q = q \otimes \hat{q} \approx \left[\frac{1}{2} \delta\theta^T, 1\right]^T\) approx. for quaternion \(\delta q\).
    \item \( \vec{q} \otimes \vec{p} = (q_4 + q_1i + q_2j + q_3k)(p_4 + p_1i + p_2j + p_3k) \) where the quaternion multiplication is defined by operation \( \otimes\).
\end{itemize}
Fig.~\ref{fig:frames} shows the coordinate frames used in this work.

\begin{figure*}[t]
  \includegraphics[width=\textwidth]{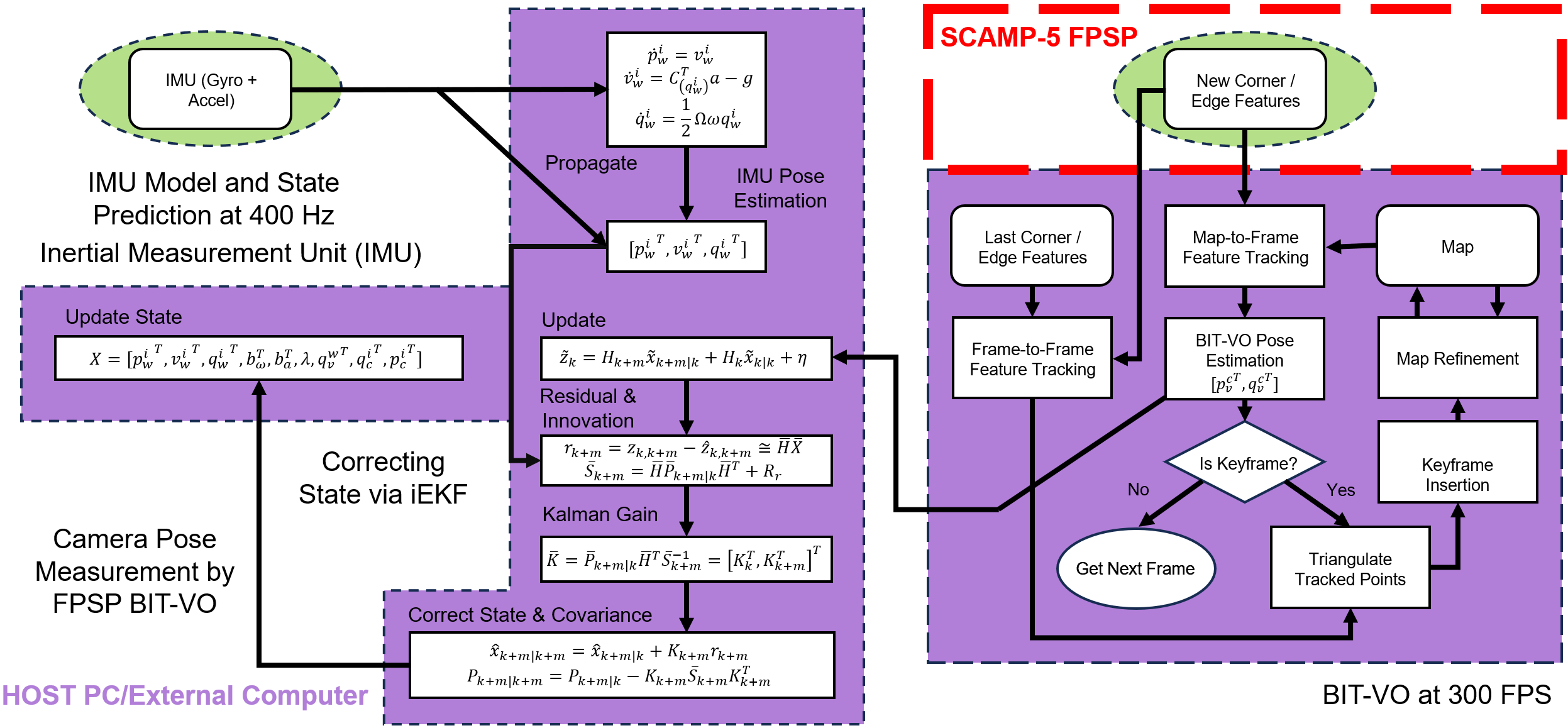}
  \vspace{-4 mm}
  \caption{Pipeline of BIT-VIO. The multi-sensor fusion is to the left. BIT-VO is to the right. From the BIT-VO algorithm~\cite{murai2020bit}, the vision sensor utilizes the SCAMP-5 FPSP, highlighted in red. New corner/edge features are detected via the FPSP, off-putting computational load by allowing some image and signal processing to be done on the chip before transferring to a PC host or other external device to be further processed.}
  \label{fig:overview}
\end{figure*}

\subsection{System Overview}
Fig.~\ref{fig:overview} demonstrates an overview of the system. The visual odometry pipeline is shown on the right, and the inertial pipeline is shown on the left. The algorithmic components of the visual odometry are mostly done on a remote host, except the corner/edge detection, which is done on the SCAMP-5 FPSP. 

\subsection{IMU Model and State Prediction}
We use iterated Extended Kalman Filter Multi-Sensor Fusion framework (iEKF-MSF)~\cite{lynen2013robust} and assume absolute IMU measurements have bias \(b_{\omega}, b_{a}\) with Gaussian noise \(n_{\omega}, n_{a}\). 
IMU measures and outputs angular velocities \(\omega\) and linear accelerations \(a\) in the IMU-frame~\cite{weiss2011real}:
\begin{align}
\omega &= \omega_{\text{meas}} - b_{\omega} - n_{\omega}\, ,\quad \dot b_{\omega} = n_{b_{\omega}}\,,  \label{eq:w_noise}\\
a &= a_{\text{meas}} - b_{a} - n_{a} \,,\quad \dot b_{a} = n_{b_{a}}\,. \label{eq:a_noise}
\end{align}

In Eq.~\eqref{eq:w_noise} and~\eqref{eq:a_noise}, the subscript ``meas" means the measured value. Terms \(\dot b_{\omega}, \dot b_{a}\) are the dynamic models of the IMU biases. 

The iEKF-MSF states \(x\) are represented in two parts: the \(x_{IMU}^{T}\) and \(x_{BIT-VO}^{T}\). The \(x_{IMU}^{T}\), which is a \color{black}16-element state\color{black}, is formed by the IMU measurements and dynamic models, as follows~\cite{weiss2011real}:
\begin{align}
x_{IMU}^{T} &= [  {p_{w}^{i}}^T, {v_{w}^{i}} ^T , {q_{w}^{i}} ^T, b_{\omega}^{T}, b_{a}^{T}]\,, \label{eq:ximu}\\
\dot{p}_{w}^{i} &= v_{w}^{i}\,, \label{eq:xpdot}\\
\dot{v}_{w}^{i} &= C^{T}_{(q_{w}^{i})} a - g\,, \label{eq:vdot}\\
\dot{q}_{w}^{i} &= \frac{1}{2} \Omega {\omega} q_{w}^{i}\,. \label{eq:qdot}
\end{align}

In Eq.~\eqref{eq:ximu}-\eqref{eq:qdot}, \({p_{w}^{i}}^{T}, {v_{w}^{i}}^{T}, {q_{w}^{i}}^{T}\) represents the translation, velocity, and quaternion rotation of the IMU w.r.t. world (or inertial frame). The dynamic models \(\dot{p}_{w}^{i}, \dot{v}_{w}^{i}, \dot{q}_{w}^{i}\) propagate the state and do so at the rate of the IMU. 

\subsection{Camera Pose Measurement by FPSP BIT-VO}

In the BIT-VO~\cite{murai2020bit}, the front-end visual processing occurs on the SCAMP-5 FPSP. FAST corner and binary edge features are detected on the chip before it is transferred to a PC host or other external device. On the host device, the visual features are further processed to obtain camera pose estimation (unscaled as the system is monocular), \(x_{BIT-VO}^{T}\), which is composed of \({p_{w}^{v}}^{T}, {q_{w}^{v}}^{T}\), i.e. position and orientation. 

BIT-VO uses a BIT-descriptor (44-bit long feature), which is created from local binary edge information around the corner features and is used to establish feature correspondences between frames. 
It differs from other binary descriptors because BIT-VO does not have access to the image intensity information. To create a BIT-descriptor, around a corner feature, BIT-VO creates a \(7 \times 7\) patch and rotates the patch to be rotationally invariant. In the \(7 \times 7\) patch, BIT-VO creates 3 rings \(r \in \{r_{1}, r_{2}, r_{3}\}\). To establish a correspondence, hamming distance between two features (as the feature is binary) is taken to measure how similar two descriptors are.
Though the BIT descriptor is rotation invariant, it is not scale invariant.

The high frame rate feature detection using SCAMP-5 FPSP simplifies the frame-to-frame and map-to-frame matching processes, as the inter-frame motion between frames is small. This allows the feature matching to be based upon simple brute-force search-and-match around a small radius (3-5 pixels) of the said features.

The map refinement and keyframe selection of the BIT-VO algorithm are similar to PTAM~\cite{4538852} and SVO~\cite{forster2016svo}. Initialization is done by the 5-point algorithm with RANSAC.

Once the 3D map points and their corresponding $k$-projected points on the image plane are found, the pose is estimated by minimizing the reprojection error:
\begin{align}
[{p_{v}^{c}}^{T},{q_{v}^{c}}^{T}] &= \argmin_{[{p_{v}^{c}}^{T},{q_{v}^{c}}^{T}]} \frac{1}{2} \sum_{i=0}^{k} \rho \left( \left\| u_{i} - \pi({T_{v}^c} \cdot {}_v {p_{i}}) \right\|^{2} \right),
\label{eq:argmineq}
\end{align}

\noindent where $\pi({T_{v}^c} \cdot {}_v {p_{i}})$ is the function projecting 3D points on the vision image plane and \(\rho(\cdot)\) is the Huber loss function, reducing the effect of outlying data. 


The \(x_{BIT-VO}^{T}\) (scaled with scale \(\lambda\)) part of the \color{black}10-element \color{black} state is defined as:
\begin{align}
x_{BIT-VO}^{T} &= [\Delta \lambda, \delta {\theta_{i}^{c}}^T, \Delta 
 {p_{w}^{v}}^{T}, \delta {\theta_{w}^{v}}^{T}]
 \label{eq:x-bit-vo-state-eq}
\end{align}

We assume BIT-VO vision sensor measurement \(z_{BIT-VO}\) has Gaussian noise in position and rotational \({n}_{p}, {n}_{q}\). The measurement model is given by,
\begin{align}
z_{BIT-VO} &= \begin{bmatrix}
p_{v}^{c} \\
q_{v}^{c}
\end{bmatrix} \\
&= \begin{bmatrix}
C_{(q_{w}^{v})} (p_{w}^{i} + C_{(q_{w}^{i})}^T p_{i}^{c}) \lambda + p_{w}^{v} + n_{p_v} \\
q_{i}^{c}\otimes q_{w}^{i} \otimes q_{w}^{v-1} + n_{q_v}
 \label{eq:z-bit-vo-state-eq}
\end{bmatrix},
\end{align}

\(p_{v}^{c}, q_{v}^{c}\) propagate the state and do so at the BIT-VO vision sensor rate, which is the rate of 300 FPS \cite{trawny2005indirect}\cite{weiss2011real}. 

\subsection{Uncertainty Propagation of FPSP BIT-VO Pose}

BIT-VO itself does not propagate an uncertainty or consist of covariance for its vision 6-DOF pose. 3D map points and correspondences are computed on the PC, where the pose is optimized by minimizing the reprojection error. Once the optimal pose $[{p_{v}^{c}}^{T},{q_{v}^{c}}^{T}]$ is found from the set, we take the pose and, using Ceres~\cite{Agarwal_Ceres_Solver_2022}, generate a $6\times 6$ covariance block for the optimized parameters based on the optimal pose. It starts with forming the Jacobian of the residual blocks with respect to $[{p_{v}^{c}}^{T},{q_{v}^{c}}^{T}]$, then the Hessian $H$ is approximated as $J^T J$, lastly with the covariance being computed as the inverse of the approximated Hessian $\Sigma = H^{-1}=(J^T J)^{-1}$. Note, here the covariance matrix is a $6\times 6$ positive definite matrix, correctly matching the system's DOF rather than the state's dimensionality (which is 7 as we have 3 parameters for the translation and 4 parameters for the quaternion).


\subsection{Correcting State via iEKF}

\begin{algorithm}[t!]
\color{black}
\caption{Correcting State via iEKF-MSF Update Process}
\label{alg:algorithm}
\begin{algorithmic}[1]
\color{black}
\STATE Build full covariance matrix $\bar{P}_{k+m|k}$.

\STATE Update is $\tilde{z}_k = H_{k+m}\tilde{X}_{k+m|k} + H_k\tilde{X}_{k|k} + \eta$, where $H$ is the measurement Jacobian found via $z_{BIT-VO}$.

\STATE Compute residual $r_{k+m} = z_{k,k+m} - \hat{z}_{k,k+m} \stackrel{\sim}{=} \bar{H}\bar{X}$.

\STATE Compute innovation $\bar{S}_{k+m} = \bar{H}\bar{P}_{k+m|k}\bar{H}^T + R_r$, where $R_r$ is covariance of measurement and $\bar{H} = [H_{k|k}, H_{k+m|k}]$.

\STATE Compute $\bar{K} = \bar{P}_{k+m|k}\bar{H}^T\bar{S}_{k+m}^{-1} = [K_k^T, K_{k+m}^T]^T$.

\STATE Correct state $\hat{x}_{k+m|k+m} = \hat{x}_{k+m|k} + K_{k+m}r_{k+m}$ and covariance $P_{k+m|k+m} = P_{k+m|k} - K_{k+m}\bar{S}_{k+m}K_{k+m}^T$.

\end{algorithmic}
\end{algorithm}

We may either assume BIT-VO vision sensor measurements as relative (as in depending between time-instants \(k\) and \(k+m\)) or absolute (e.g. IMU or GPS measurements). If it is a relative measurement, see Alg.~\ref{alg:algorithm}. Otherwise, if absolute, the algorithm remains the same, but the updates must occur on \(k\) not \(k+m\). 

\section{Experimental Results}
\label{sec:results}

This section presents the experimental results. First, we explain the experimental setup and the calibration of the system, then we discuss the experimental results. We evaluate the system on eight different trajectories (labelled A-H).

\begin{figure}[t]
\centerline{\includegraphics[width = 0.3 \textwidth]{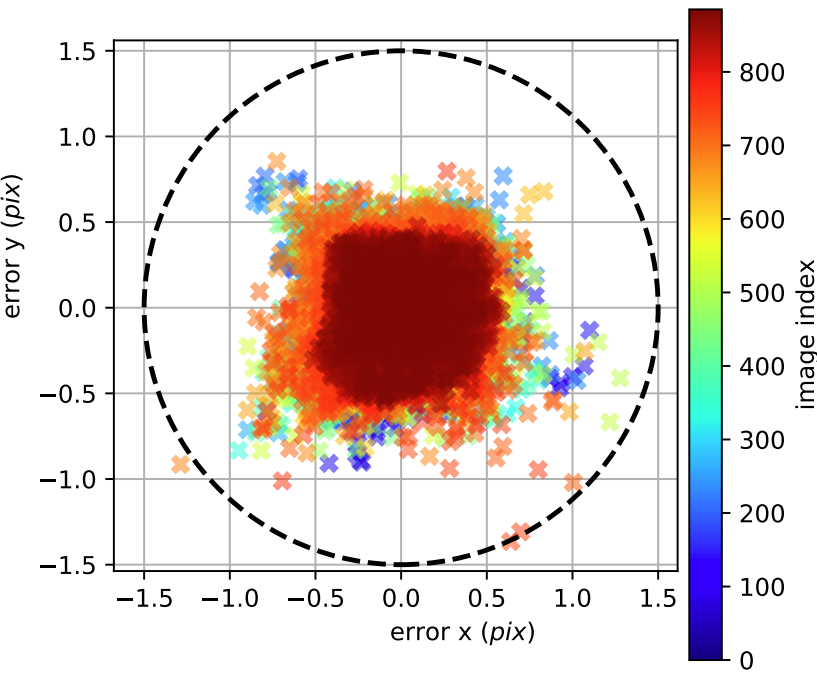}}
 \vspace{-3 mm}
\caption{\color{black}FPSP SCAMP-5 reprojection error on a $256\times 256$ focal-plane imaging output with a less than one-pixel error (within the dashed black).\color{black}}
\label{fig:cam0error}
\end{figure}

\subsection{Experimental Setup}
\color{black}
The proposed BIT-VIO algorithm is tested at \color{black} 300 FPS, running BIT-VO on a SCAMP-5 FPSP device. Additionally, we attach an Intel D435i RealSense Camera to provide IMU measurements at 400 Hz. \color{black} 

\color{black}Evaluations are done against ground-truth data from a Vicon motion capture system, which consisted of 14 cameras calibrated and time-synced. As both BIT-VO and BIT-VIO assume a fast frame rate, hence small inter-frame motion, we cannot evaluate our method using a standard benchmark dataset for direct comparison with other methods. Hence, BIT-VIO is evaluated on eight real-world trajectories against BIT-VO. These trajectories are designed to mimic practical applications and are a compilation of circular, straight, curved, and zigzag trajectories. 
The recorded trajectories are aligned and scaled to the ground-truth trajectory as our setup is monocular. We measure the Absolute Trajectory Error (ATE)~\cite{lu1997globally} and report the Root Mean Squared Error (RMSE)~\cite{sturm2012benchmark} together with the median to evaluate the accuracy against the ground-truth.
BIT-VO and BIT-VIO use a host device to perform the visual odometry backend, and for the host device, we use an external laptop with 13th Gen Intel Core i7-12700 CPU.

\subsection{Sensor Calibration}
We use Kalibr \cite{furgale2013unified} to perform the calibration between SCAMP-5 and the IMU.
For the extrinsic calibration, we obtain \color{black}\(p_i^{c} = (0.006m,0.04m,0.07m)\) \color{black} with respect to the IMU-frame.
To calibrate the IMU intrinsics: acceleration, gyroscopic, and bias noises, \(n_a, n_\omega, b_a, b_\omega\), we conducted a calibration using the Allan variance method~\cite{furgale2012continuous},~\cite{furgale2013unified},~\cite{rehder2016extending},~\cite{zhang2002clock}.
We achieve an accelerometer noise density and random walk: $[0.001865 m/s^2/\sqrt{Hz}, 0.002m/s^3/\sqrt{Hz}]$ and gyroscope noise density and random walk $[0.001865m/s^2/\sqrt{Hz}, 4\times 10^{-6}m/s^3/\sqrt{Hz}]$. The IMU gyroscope bias intrinsics error estimates are within the 3-$\sigma$ error bound. To calibrate the camera intrinsics we first estimated the focal length, camera center $[fx, fy, cx, cy]$, and distortion coefficients, and then optimized the intrinsics by optimization on a radtan lens~\cite{kannala2006generic},~\cite{maye2013selfsupervised}. We achieve a focal length: $[257.27, 258.00]$ pixels and principal point: $[127.44, 128.17]$ pixels, with a less than one pixel error as shown in Fig.~\ref{fig:cam0error}.

\begin{table}[t]
\caption{\color{black}ATE comparison of BIT-VIO and BIT-VO. The lower ATE is emphasized in bold.}
\centering
\footnotesize 
\setlength{\tabcolsep}{2pt} 
\renewcommand{\arraystretch}{1.4} 
\makebox[0.5\textwidth][c]{%
\begin{tabular}{ccccc}
\hline
Traj. & Type & BIT-VO ATE (m) & BIT-VIO ATE (m) & Length (m) \\ 
\hline

\multirow{2}{*}{A} & RMSE: & 0.215732 & \textcolor{black}{\textbf{0.167631}} & 2.1 \\
 & median: & 0.170214 & \textcolor{black}{\textbf{0.152106}} & \\
\hline


\multirow{2}{*}{B} & RMSE: & 0.134617 & \textcolor{black}{\textbf{0.12071}} & 2.6 \\
 & median: & 0.119079 & \textcolor{black}{\textbf{0.111856}} & \\
\hline
\multirow{2}{*}{C} & RMSE: & 0.094479 & \textcolor{black}{\textbf{0.086911}} & 1.7 \\
 & median: & 0.07561 & \textcolor{black}{\textbf{0.068756}} & \\
\hline
\multirow{2}{*}{D} & RMSE: & 0.175323 & \textcolor{black}{\textbf{0.153335}} & 2.12 \\
 & median: & 0.174444 & \textcolor{black}{\textbf{0.140952}} & \\
\hline
\multirow{2}{*}{E} & RMSE: & 0.206866 & \textcolor{black}{\textbf{0.195263}} & 2.4 \\
 & median: & 0.15714 & \textcolor{black}{\textbf{0.149103}} & \\
\hline


\multirow{2}{*}{F} & RMSE: & \textbf{0.134361} & 0.134328 & 2.01 \\
 & median: & \textbf{0.116587} & 0.124618 & \\
\hline

\multirow{2}{*}{G} & RMSE: & 0.132624 & \textcolor{black}{\textbf{0.10535}} & 1.8 \\
 & median: & 0.125924 & \textcolor{black}{\textbf{0.095664}} & \\
\hline

\multirow{2}{*}{H} & RMSE: & 0.10864 & \textcolor{black}{\textbf{0.104366}} & 2.55 \\
 & median: & 0.089689 & \textcolor{black}{\textbf{0.08788}} & \\
\hline



\end{tabular}\label{tb:table1}
}
\end{table}

\begin{figure*}[t]
\begin{center}
  \includegraphics[width=\textwidth]{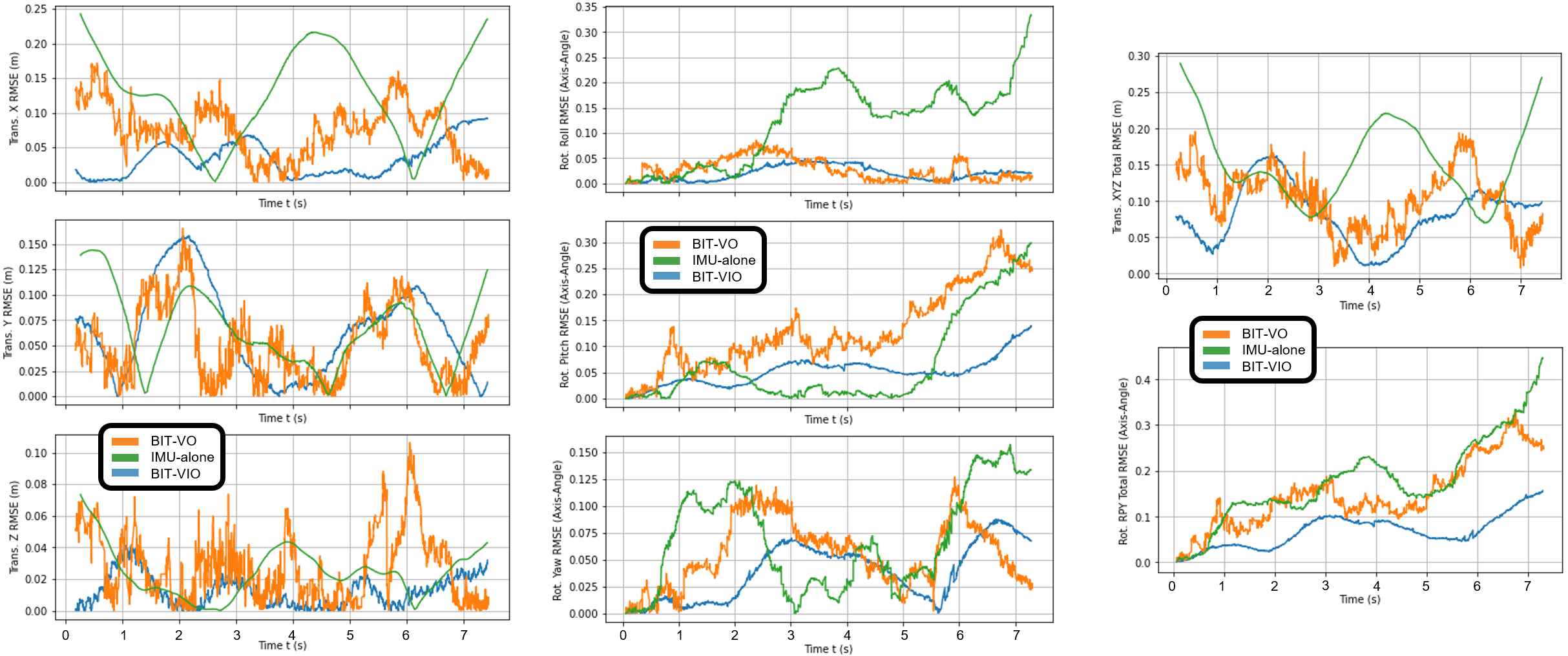}
  \vspace{-3 mm}
  \caption{Plots of the estimated translational RMSE (left) and rotational RMSE (middle) for Traj. G from Table~\ref{tb:table1}. To the very right is the total translational RSME (top) and total rotational RMSE (bottom). 
  For both translation and rotation, BIT-VIO is much closer and smoother to ground-truth data than IMU-alone and BIT-VO. The drift of the IMU-alone is very evident, as well as the high-frequency noise of BIT-VO.
  }
  \label{fig:trajL}
\end{center}
\end{figure*}

\begin{figure*}[t]
\begin{center}
  \includegraphics[width=0.8\textwidth]{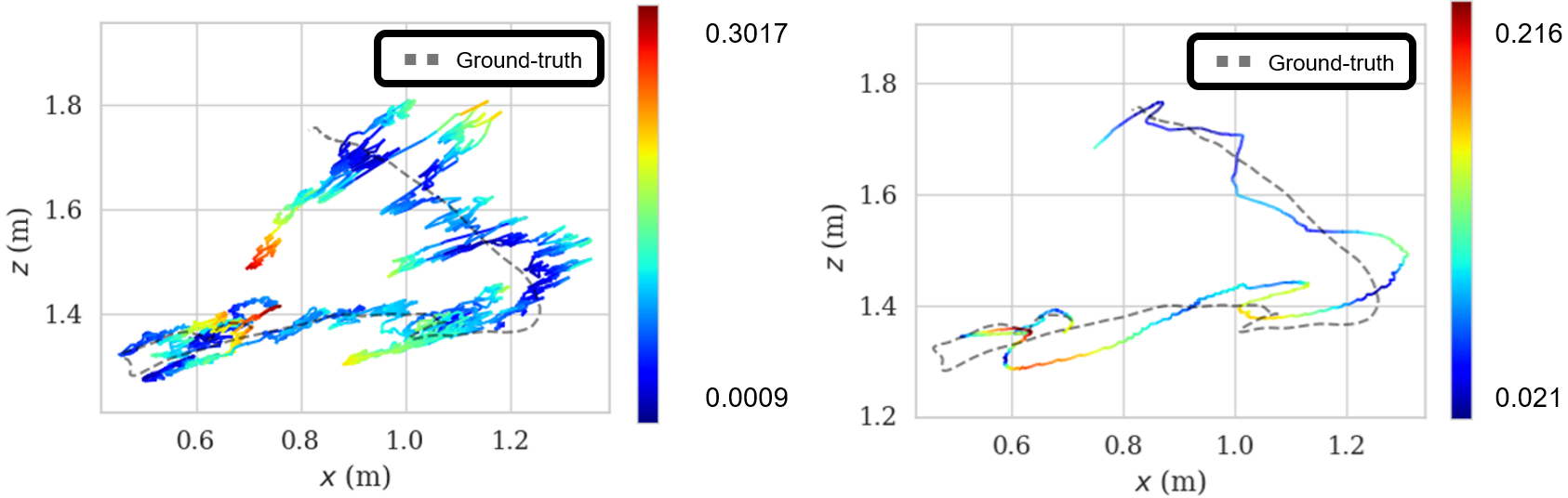}
   \vspace{-3 mm}
  \caption{Projection of \color{black}Traj. H from Table~\ref{tb:table1} onto xz-plane for BIT-VO (left) and BIT-VIO (right). We observe that BIT-VO suffers from high-frequency noise when compared to BIT-VIO's estimates, although the overall RMSE ATE are similar.\color{black}}
  \label{fig:error-mapping}
\end{center}
\end{figure*}


\subsection{Accuracy and Robustness}


\color{black} As shown in Table~\ref{tb:table1}, when incorporating an IMU, the state generally enhances its estimation with a more accurate trajectory, showcasing lower RMSE and median closer to the ground-truth values. Traj. A and B are circular and curved, Traj. C is straight, and the rest are combinations of all with zigzag. The case of Traj. G in Fig.~\ref{fig:trajL}, shows that IMU-alone accumulates error and drifts away from ground-truth data, as shown in the large translational, rotational RMSE. In fact, it has the largest RMSE compared to BIT-VO and BIT-VIO. The BIT-VIO algorithm fixes this IMU error drift, using the BIT-VO update to align it closer to ground-truth data, hence why its RMSE is the lowest of the three. This is true in both the translational and rotational context, where we can see that in the left and middle plots of Fig.~\ref{fig:trajL}, BIT-VIO RMSE generally resides much more below both IMU and BIT-VO. To add, the BIT-VIO algorithm deals well with fast, hostile motions, covering the main limitation of the prior work BIT-VO with the high-frequency noise on its predicted trajectories. Through all plots in Fig.~\ref{fig:trajL}, BIT-VO maintains its noise. BIT-VIO not only maintains itself closer to ground-truth trajectory but also does well to track smoothly with less noise, especially in more violent, quick, hostile motions.

In Fig.~\ref{fig:error-mapping}, we can see projecting the trajectory error for both BIT-VO and BIT-VIO on ground-truth Traj. H, onto an xz-plane, qualitatively gives us further insight into the magnitude of the high-frequency error present in BIT-VO and how much is removed by BIT-VIO. Not only the trajectory estimated by BIT-VIO is smoother, but it is also closer to the ground-truth trajectory. 



 \color{black}
\section{Conclusion}
\label{sec:conc}

We have presented BIT-VIO, the first-ever 6-Degrees of Freedom (6-DOF) Visual Inertial Odometry (VIO) algorithm, which utilizes the advantages of the SCAMP-5 FPSP for vision-IMU-fused state estimation. BIT-VIO operates and corrects by loosely-coupled sensor-fusion iterated Extended Kalman Filter (iEKF) at 300 FPS with an IMU at 400 Hz. We evaluate BIT-VIO against BIT-VO and demonstrate improvements in ATE across many trajectories. Moreover, the high-frequency noise evident in BIT-VO is effectively filtered out, resulting in a smoother estimated trajectory. 

In the future, we plan to take the next steps toward a tightly-coupled VIO approach using the SCAMP-5 FPSP. 

\color{black}

\section{Acknowledgements}

\color{black}We would like to thank Ali Babaei and Abrar Ahsan for their early help with the SCAMP-5. \color{black}This research is supported by Natural Sciences and Engineering Research Council of Canada (NSERC). We would like to thank Piotr Dudek, Stephen J. Carey, and Jianing Chen at the University of Manchester for kindly providing access to SCAMP-5.

\bibliographystyle{plain}
\bibliography{main.bib}

\end{document}